\documentclass[runningheads]{llncs}
\usepackage{graphicx}
\usepackage{amsmath,amssymb} 
\usepackage{color}

\usepackage{booktabs}
\newcommand{\eg}{\textit{e.g.}}
\newcommand{\ie}{\textit{i.e.}}
\newcommand{\etal}{\textit{et al.}}

\usepackage{array}
\newcolumntype{L}[1]{>{\raggedright\let\newline\\\arraybackslash\hspace{0pt}}m{#1}}
\newcolumntype{C}[1]{>{\centering\let\newline\\\arraybackslash\hspace{0pt}}m{#1}}
\newcolumntype{R}[1]{>{\raggedleft\let\newline\\\arraybackslash\hspace{0pt}}m{#1}}

\begin{document}

\newcommand{\point}{
    \raise0.7ex\hbox{.}
    }


\pagestyle{headings}

\mainmatter

\title{Video Summarization using Deep Semantic Features} 

\titlerunning{Video Summarization using Deep Semantic Features} 

\authorrunning{M. Otani \etal} 

\author{Mayu Otani\inst{1} \and Yuta Nakashima\inst{1} \and Esa Rahtu\inst{2} \and \\Janne Heikkil\"a\inst{2} \and Naokazu Yokoya\inst{1}}

\institute{{Graduate School of Information Science, Nara Institute of Science and Technology\\
	\email{ \{otani.mayu.ob9,n-yuta,yokoya\}@is.naist.jp}}\and
    {Center for Machine Vision and Signal Analysis, University of Oulu\\
	\email{ \{erahtu,jth\}@ee.oulu.fi}}
}

\maketitle

\begin{abstract}
This paper presents a video summarization technique for an Internet video to provide a quick way to overview its content. This is a challenging problem because finding important or informative parts of the original video requires to understand its content. Furthermore the content of Internet videos is very diverse, ranging from home videos to documentaries, which makes video summarization much more tough as prior knowledge is almost not available. To tackle this problem, we propose to use deep video features that can encode various levels of content semantics, including objects, actions, and scenes, improving the efficiency of standard video summarization techniques. For this, we design a deep neural network that maps videos as well as descriptions to a common semantic space and jointly trained it with associated pairs of videos and descriptions. To generate a video summary, we extract the deep features from each segment of the original video and apply a clustering-based summarization technique to them. We evaluate our video summaries using the SumMe dataset as well as baseline approaches. The results demonstrated the advantages of incorporating our deep semantic features in a video summarization technique.
\end{abstract}

\section{Introduction}
With the proliferation of devices for capturing and watching videos, video hosting services have gained an enormous number of users. According to \cite{youtube_stat} for example, almost one third of the people online use YouTube to upload or review videos. This increasing popularity of Internet videos has accelerated the demand for efficient video retrieval. Current video retrieval engines usually rely on various types of metadata, including title, user tags, descriptions, and thumbnails, to find videos, which is usually given by video owners. However, such metadata may not be very descriptive to represent the entire content of a video. Moreover, titles and tags are completely up to video owners and so their semantic granularity can vary video by video, or such metadata can even be irrelevant to the content. Consequently users need to review retrieved videos, at least partially, to get rough ideas on their content. 

One potential remedy for this comprehensibility problem in video retrieval results is to adopt video summarization, which generates a compact representation of a given video. By providing such summaries as video retrieval results, the users can easily and quickly find desired videos. Video summarization has been one of the major areas in the computer vision and multimedia fields, and a wide range of techniques have been proposed for various goals. Among them, ideal video summarization tailored for the comprehensibility problem should include video content that is essential to tell the story in the entire video. At the same time, it also needs to avoid inclusion of semantically unimportant or redundant content. 

To this end, many existing approaches for video summarization extract short video segments based on a variety of criteria that are designed to find essential parts with small redundancy. Examples of such approaches include sampling some exemplars from a set of video segments based on visual features \cite{Gong2000,Gong2014} and detecting occurrences of unseen content \cite{Zhao2014}. These approaches mostly rely on low-level visual features, \eg, color histogram, SIFT \cite{Lowe2004}, and HOG \cite{Dalal2005}, which are usually deemed far from the semantics. Some recent approaches utilize higher-level features including objects and identities of people. Their results are promising, but they cannot handle various concepts except a predefined set of concepts, while an Internet video consists of various levels of semantic concepts, such as objects, actions, and scenes. Enumerating all possible concepts as well as designing concept detectors are almost infeasible, which makes video summarization challenging. 

\begin{figure}[t!]
    \centering
    \includegraphics[width=0.9\textwidth]{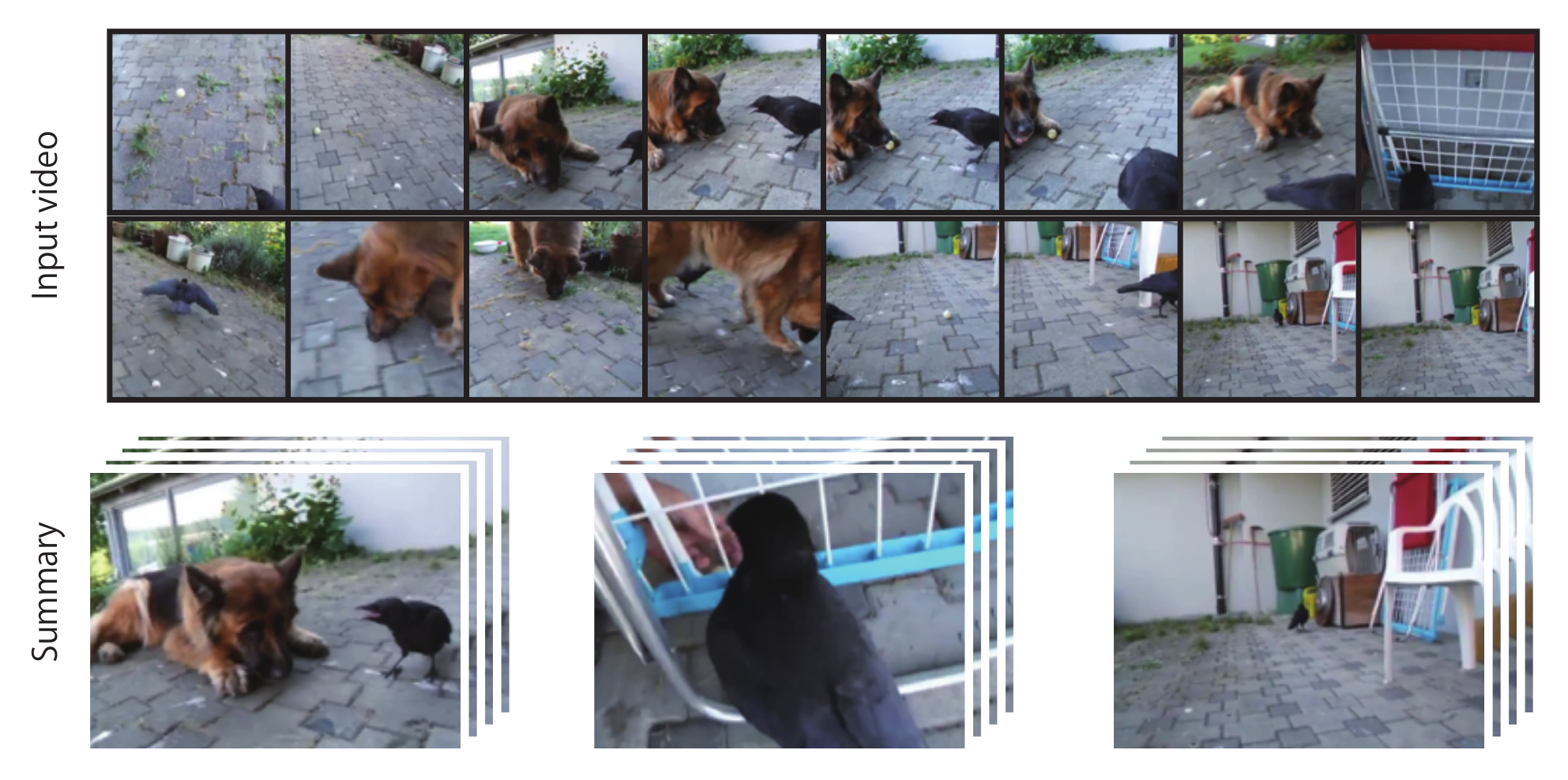}
    \caption{An example of an input video and a generated video summary. The same content (\ie, the dog) repeatedly appears in the input video in different appearances or background, which may be semantically redundant. Our video summary successfully reduces such redundant video segments, thanks to our deep features encoding higher-level semantics.}
    \label{fig:video_summary_overview}
\end{figure}

This paper presents a novel approach for video summarization. Our approach enjoys recent advent of deep neural networks (DNNs). Our approach segments the original videos into short video segments, for each of which we calculate deep features in a high-dimensional, continuous semantic space using a DNN. We then sample a subset of video segments such that the sampled segments are semantically representative of the entire video content and are not redundant. For sampling such segments, we define an objective function that evaluates representativeness and redundancy of sampled segments. After sampling video segments, we simply concatenate them in the temporal order to generate a video summary (Fig.~\ref{fig:video_summary_overview}).

To capture various levels of semantics in the original video, deep features play the most important role. Several types of deep features have been proposed recently using convlutional neural networks (CNNs) \cite{Yao2015,Donahue2014}. These deep features are basically trained for a certain classification task, which predicts class labels of a certain domain, such as objects and actions. Being different from these deep features, our deep features need to encode a diversity of concepts to handle a wide range of Internet video contents. To obtain such deep features, we design a DNN to map videos and descriptions to the semantic space and train it with a dataset consisting of videos and their associated descriptions. Such a dataset contains descriptions like ``a man is playing the guitar on stage,'' which includes various levels of semantic concepts, such as objects (``man'', ``guitar''), actions (``play''), and a scene (``on stage''). Our DNN is jointly trained using such a dataset so that a pair of a video and its associated sentence gives a smaller Euclidean distance in the semantic space. We use this DNN to obtain our deep features; therefore, our deep features well capture various levels of semantic concepts.

The contribution of this work can be summarized as follows:
\begin{itemize}
\item We develop deep features for representing an original input video. In order to obtain features that capture higher level semantics and are well generalized to various concepts, our approach learns video features using their associated descriptions. By jointly training the DNN using videos and descriptions in recently released large-scale video-description dataset \cite{Xu2016}, we obtain deep features capable of encoding sentence-level semantics.
\item We leverage the deep features for generating a video summary. To the best of our knowledge, this is the first attempt to use jointly trained deep features for the video summarization task.
\item We represent a video using deep features in a semantic space, which can be a powerful tool for various tasks like video description generation and video retrieval.
\item We quantitatively demonstrate that our deep features benefit the video summarization task, comparing ours to deep features extracted using VGG \cite{Simonyan2015}.
\end{itemize}

\section{Related Work}

\subsubsection{Video Summarization.}
The difficulty in video summarization lies in the definition of ``important'' video segments to be included in a summary and their extraction. At the early stage of video summarization research, most approaches focus on a certain genre of videos. For example, the importance of a video segment in broadcasting sports program may be easily defined based on the event happening in that segment according to the rules of the sports \cite{Babaguchi2004}. Furthermore, a game of some sports (\eg, baseball and American football) has a specific structure that can facilitate important segment extraction. Similarly, characters that appear in movies are also used as domain knowledge \cite{Sang2010}. For these domains, various types of metadata (\eg, a textual record of scoring in a game, movie scripts, and closed captions) help to generate video summaries \cite{Babaguchi2004,Sang2010,Evangelopoulos2013}. Egocentric videos are another interesting example of video domains, for which a video summarization approach using a certain set of predefined objects as a type of domain knowledge has been proposed \cite{Lu2013}. More recent approaches in this direction adopt supervised learning techniques to embody domain knowledge. For example, Potapov \etal~\cite{Potapov2014} proposed to summarize a video focusing on a specific event and used an event classifier's confidence score as the importance of a video segment. Such approaches, however, are almost impossible to generalize to other genres because they heavily depend on domain knowledge.

In the last few years, video summarization has been addressed in an unsupervised fashion or without using any domain knowledge. Such approaches introduce the importance of video segment by using various types of criteria and cast video summarization into an optimization problem involving these criteria. Yang \etal~\cite{Yang2015} proposed to utilize an auto-encoder, in which its encoder converts an input video's features into a more compact one, and the decoder then reconstructs the input. The auto-encoder is trained with Internet videos in the same topic. According to the intuition that the decoder can well reconstruct features from videos with frequently appearing content, they assess the segment importance based on the reconstruction errors. Another innovative approach was presented by Zhao \etal, which finds a video summary that well reconstructs the rest of the original video. The diversity of segments included in a video summary is an important criterion and many approaches use various definitions of the diversity \cite{Gong2014,Xu2015,Tschiatschek2014}.

These approaches used various criteria in the objective function, but their contributions have been determined heuristically. Gygli \etal~added some supervised flavor to these approaches for learning each criterion's weight  \cite{Gygli2014,Gygli2015}. One major problem of these approaches is that such datasets do not scale because manually creating good video summaries is cumbersome for people. 

Canonical views of visual concepts can be an indicator of important video segments, and several existing work uses this intuition for generating a video summary \cite{Song2015,Khosla2013,Chu2015}. These approaches basically find canonical views in a given video, assuming that results of image or video retrieval using the video's title or keywords as query contain canonical views. Although a group of images or videos retrieved for the given video can effectively predict the importance of video segments, retrieving these images/videos for every input video is expensive and can be difficult because there are only a few relevant images/videos for rare concepts.

For the goal of summarizing Internet videos, we employ a simple algorithm for segment extraction. This is very different from the above approaches that use a sophisticated segment extraction method relying on low-level visual features with manually created video summaries or topic specific data. Due to the dependency of low-level visual features, they do not distinguish semantically identical concepts with different appearances caused by different viewpoints or lighting conditions, and consequently result in semantically redundant video summaries. Instead of designing such a sophisticated algorithm, we focus on designing good features to represent the original video with richer semantics, which can be viewed as the counterpart of sentences' semantics.

\subsubsection{Representation Learning.}
Recent research efforts on CNNs have revealed that the activations of a higher layer of a CNN can be powerful visual features \cite{Donahue2014,Wang2015}, and CNN-based image/video representations have been explored for various tasks including classification \cite{Donahue2014}, image/video retrieval \cite{Frome2013,Xu2015}, and video summarization \cite{Gygli2015,Yang2015}. Some approaches learn deep features or metrics between a pair of inputs, possibly in different modalities, using a Siamese network  \cite{Chopra2005,Lin2015}. Kiros \etal~\cite{Kiros2015} proposed to retrieve image using sentence queries and vice versa by mapping images and sentences into a common semantic space. For doing this, they jointly trained the mappings using video-description pairs and the contrastive loss such that positive pairs (\ie, an image and a relevant sentence) and negative pairs (\ie, an image and a randomly selected irrelevant sentence) give smaller and larger Euclidean distances in the semantic space, respectively. 

Inspired by Kiros \etal's work, we develop a common semantic space, which is also jointly trained with pairs of videos and associated sentences (or descriptions). With this joint training, our deep features are expected to encode sentence level semantics, rather than word-or object-level ones. Such deep semantic features can boost the performance of a standard algorithm for important video segment extraction, \ie, clustering-based one, empowering them to cope with higher-level semantics.

\section{Approach}

\begin{figure}[t!]
    \centering
    \includegraphics[width=\textwidth]{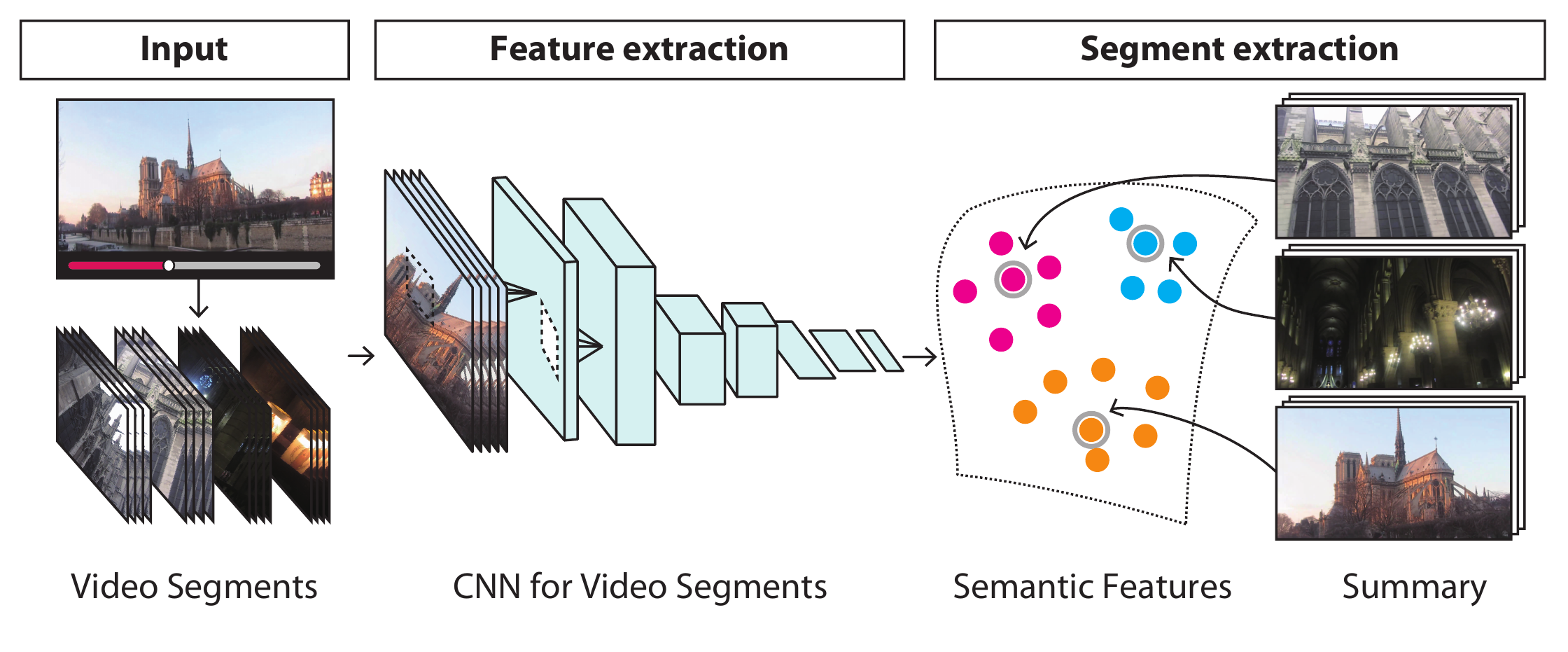}
    \caption{Our approach for video summarization using deep semantic features. We extract uniform length video segments from an input video. The segments are fed to a CNN for feature extraction and mapped to points in a semantic space. We generate a video summary by sampling video segments that correspond to cluster centers in the semantic space.}
    \label{fig:overview}    
\end{figure}

\begin{figure}[t!]
    \centering
    \includegraphics[width=\textwidth]{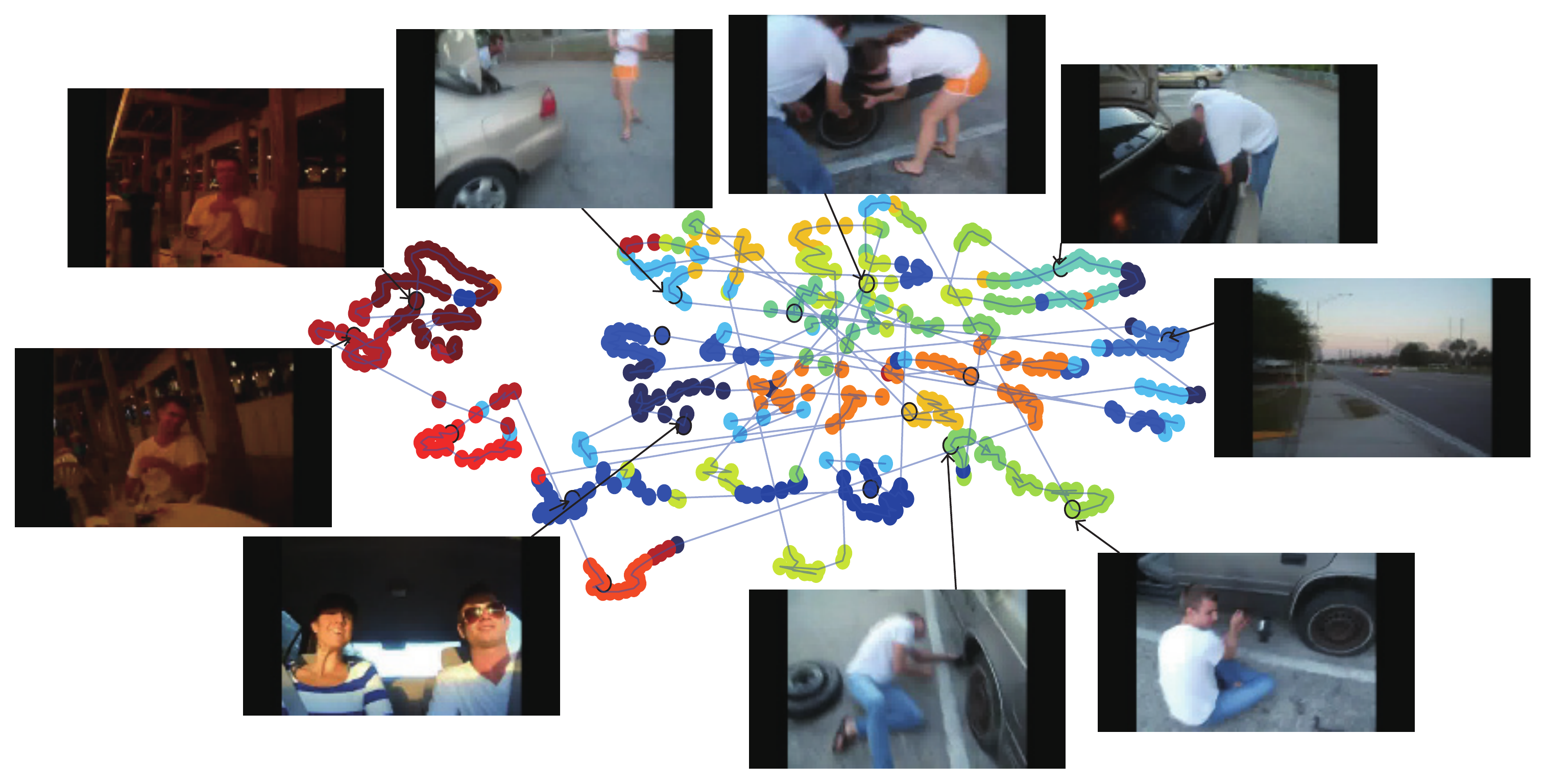}
    \caption{A two-dimensional plot of our deep features calculated from a video, where we reduce the deep features' dimensionality with t-SNE \cite{VanderMaaten2008}. Some deep features are represented by the corresponding video segments' keyframes, and the edges connecting deep features represent temporal adjacency of video segments. The colors of deep features indicate clusters obtained by k-means, \ie, points with the same color belong to the same cluster.}
    \label{fig:trajectory}
\end{figure}

Figure \ref{fig:overview} shows an overview of our approach for video summarization. We first extract uniform length video segments from the input video in a temporal sliding window manner and compute their deep semantic features using a trained DNN. Inspired by \cite{Daniel}, we represent the input video as a sequence of deep features in the semantic space, each of which corresponds to a video segment, as shown in Fig.~\ref{fig:trajectory}. This representation can encode the semantic transition of the video and thus can be useful for various tasks including video retrieval, video description generation, etc. In Fig.~\ref{fig:trajectory}, some clusters can be observed, each of which are expected to contain semantically similar video segments. Based on this assumption, our approach picks out a subset of video segments by optimizing an objective function involving the representativeness of the subset.

\begin{figure}[t!]
    \centering
    \includegraphics[width=0.6\textwidth]{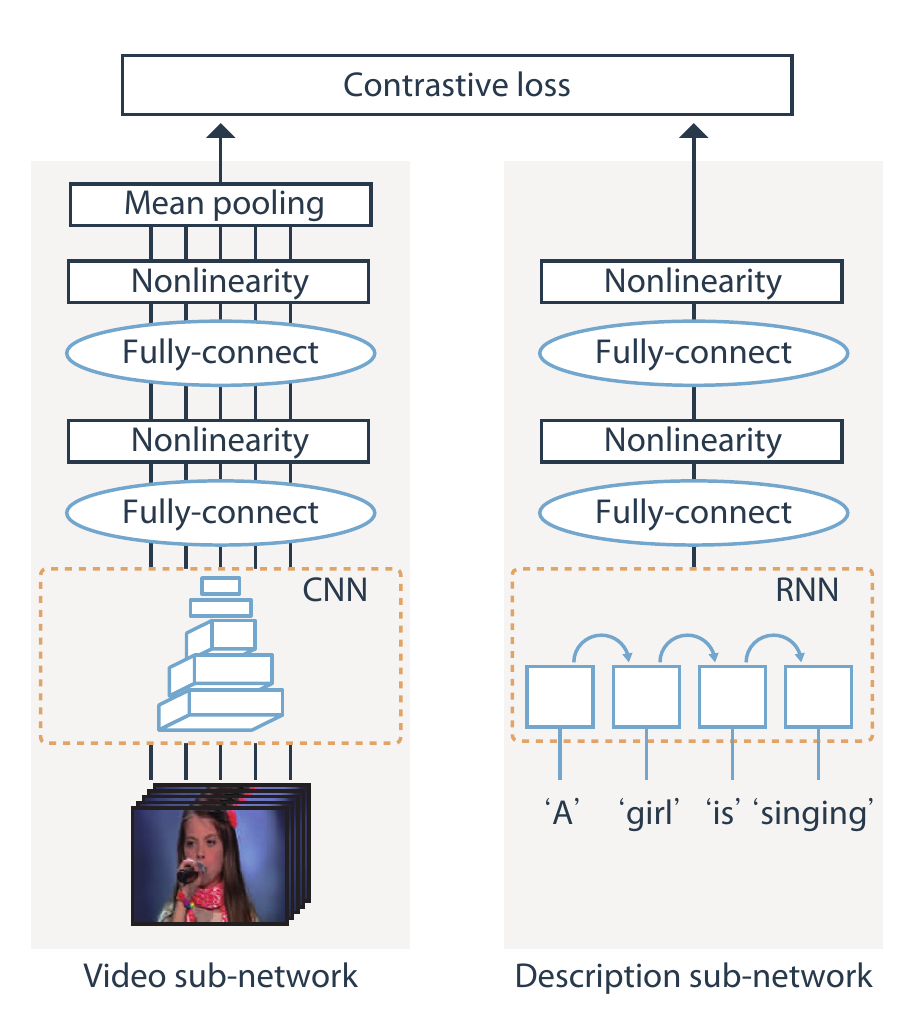}
    \caption{The network architecture. Video segments and descriptions are encoded into vectors in the same size. Both sub-network for videos and descriptions are trained jointly by minimizing the contrastive loss. }
    \label{fig:training}    
\end{figure}

The efficiency of the deep features is crucial in our approach. To obtain good deep features that can capture higher-level semantics, we use the DNN shown in Fig.~\ref{fig:training}, consisting of two sub-networks to map a video and a description to a common semantic space and jointly train them using a large-scale dataset of videos and their associated descriptions (a sentence). The video sub-network basically is a CNN, and the sentence sub-network is a recurrent neural network (RNN) with some additional layers. We use the contrastive loss function \cite{Chopra2005} for training, which tries to bring a video and its associated description (a positive pair) closer (\ie, a small Euclidean distance in the semantic space) while a video and a randomly sampled irrelevant description (a negative pair) farther. Being different from other visual features using a CNN trained to predict labels of a certain domain \cite{Donahue2014,Yao2015}, our deep features are trained with sentences. Consequently, they are expected to contain sentence-level semantics, including objects, actions, and scenes.

\subsection{Learning Deep Features}
\label{sec:feature}

To cope with higher-level semantics, we jointly train the DNN shown in Fig.~\ref{fig:training} with pairs of videos and sentences, and we use its video sub-network for extracting deep features. The video sub-network is a modified version of VGG \cite{Simonyan2015}, which is renowned for a good classification performance. In our video sub-network, VGG's classification (``fc8'') layer is replaced with two fully-connected layers with hyperbolic tangent (tanh) nonlinearity, which is followed by  a mean pooling layer to fuse different frames in a video segment.
Let $V = \{ v_i | i = 1, \ldots , M\}$ be a video segment, where $v_i$ represents frame $i$. We feed the frames to the video sub-network and compute a video representation $X \in \mathbb{R}^d$.

For the sentence sub-network, we use skip-thought vector by Kiros \etal~\cite{Kiros2015}, which encodes a sentence into 4800-dimensional vectors with an RNN. Similarly to the video sub-network, we introduce two fully-connected layers with tanh nonlinearity (but without a mean pooling layer) as in Fig.~\ref{fig:training} to calculate a sentence representation $Y \in \mathbb{R}^d$ from a sentence $S$. 

\begin{figure}[t!]
    \centering
    \includegraphics[width=\textwidth]{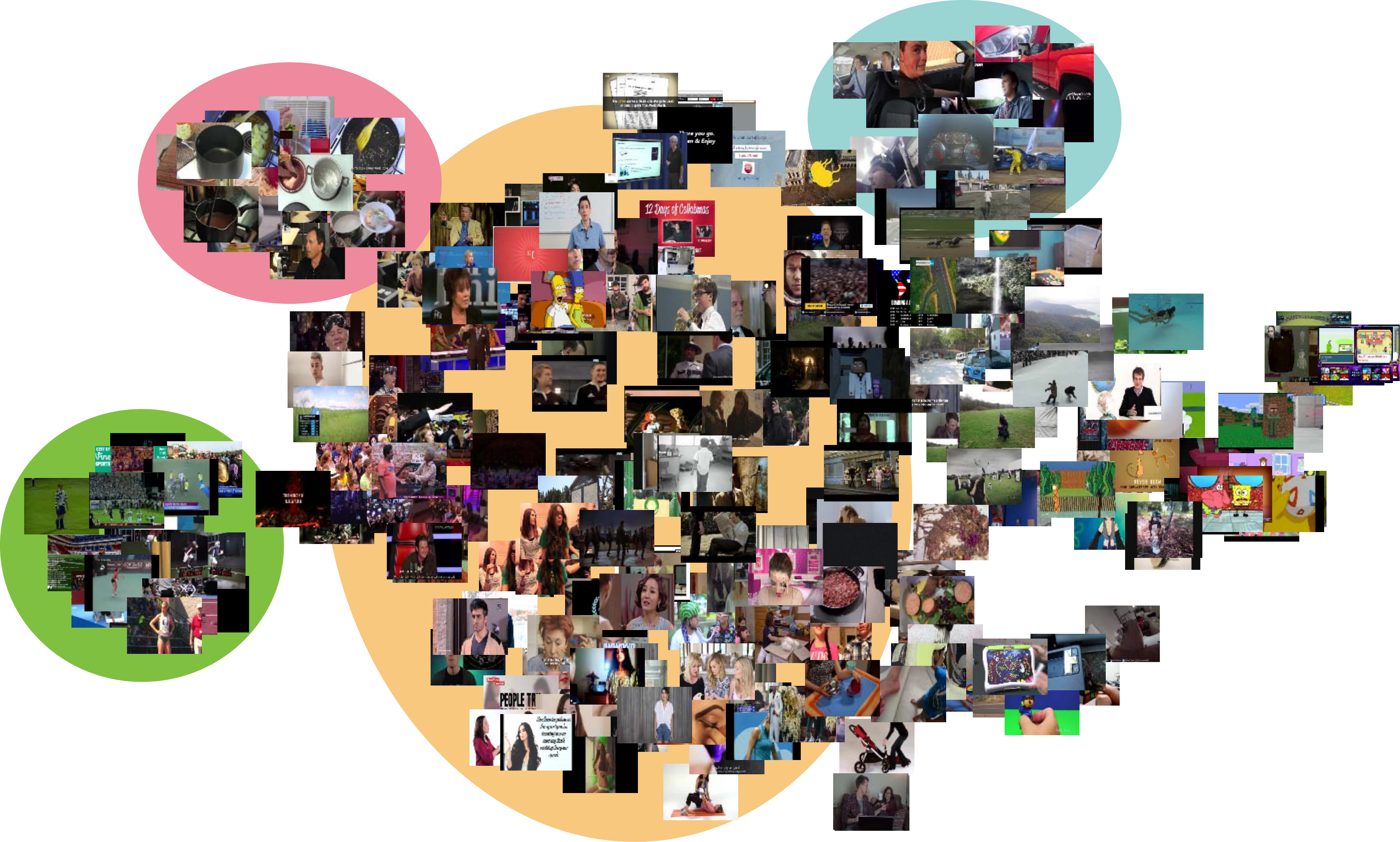}
    \caption{Two-dimensional deep feature embedding with keyframes of corresponding videos, where the feature dimensionality is reduced with t-SNE. The videos located on each colored ellipsis show similar content, \eg, cars and driving people (blue), sports (green), talking people (orange), and cooking (pink).}
    \label{fig:2d_plot}    
\end{figure}

For training these sub-networks jointly, we use a video-description dataset (\eg, \cite{Xu2016}). We sample positive and negative pairs, where a positive pair consists of a video segment and its associated description, and a negative pair consists of a video and a randomly sampled irrelevant description. Our DNN is trained with the contrastive loss \cite{Chopra2005}, which is defined using extracted features $(X_n, Y_n)$ for the $n$-th video and description pair as:
\begin{equation}
    \mathrm{loss}(X_n, Y_n) = t_n d(X_n, Y_n) + (1-t_n) \max (0, \alpha - d(X_n, Y_n)),
\end{equation}
where $d(X_n, Y_n)$ is the squared Euclidean distance between $X_n$ and $Y_n$ in the semantic space, and $t_n=1$ if pair $(X_n, Y_n)$ is positive, and $t_n= 0$, otherwise. This loss encourages associated video segment and description to have smaller Euclidean distance in the semantic space, and irrelevant ones to have larger distance.
$\alpha$ is a hyperparameter to penalizes irrelevant video segment and description pairs whose Euclidean distance is smaller than $\alpha$. In our approach, we compute Euclidean distance of positive pairs with initial DNNs before training and employ the largest distance among them as $\alpha$.
This enable most pairs to be used to update the parameters at the begining of the training.
Our DNNs for videos and descriptions can be optimized using the backpropagation technique.

Figure \ref{fig:2d_plot} shows a 2D plot of learned deep features, in which the dimensionality of the semantic space is reduced using t-SNE \cite{VanderMaaten2008} and a keyframe of each video segment is placed at the corresponding position. This plot demonstrates that our deep neural net successfully locates semantically relevant videos at closer points. For example, the group of videos around the upper left area (pink) contains cooking videos, and another group on the lower left (green) shows various sports videos.
For video summarization, we use the deep features to represent a video segment.

\subsection{Generating Video Summary}

Figure \ref{fig:trajectory} shows a two-dimensional plot of deep features from a video, whose dimensionality is reduced again using t-SNE. This example illustrates that a standard method for video summarization, \eg, based on clustering, works well because, thanks to our deep features, video segments with a similar content are concentrated in the semantic space. From this observation, we generate a video summary given an input video by solving the $k$-medoids problem \cite{Gygli2015}.

In the $k$-medoids problem, we find a subset $\mathcal{S} = \{S_k | k=1, \ldots, K\}$ of video segments, which are cluster centers that minimize the sum of the Euclidean distance of all video segments to their nearest cluster centers $S_k \in \mathcal{S}$ and $K$ is a given parameter to determine the length of the video summary. Letting $\mathcal{X} = \{X_j|j = 1, \dots, L\}$ be a set of deep features extracted from all video segments in the input video, $k$-medoids finds a subset $\mathcal{S} \subset \mathcal{X}$, that minimizes the objective function defined as:
\begin{equation}
    F(\mathcal{S}) = \sum_{X \in \mathcal{X}} \mathop{ \min }_{S \in \mathcal{S}} \| X - S \|^2_2. \label{eq:vidsum}
\end{equation}
The optimal subset 
\begin{equation}
\mathcal{S}^* = \mathop{\mathrm{arg min}}_{S} F(\mathcal{S})
\end{equation} 
includes the most representative segments in clusters. As shown in Fig.~\ref{fig:2d_plot}, our video sub-network maps segments with similar semantics to closer points in the semantic space; therefore we can expect that the segments in a cluster have semantically similar content and subset $\mathcal{S}^*$ consequently includes most representative and diverse video segments. The segments in $\mathcal{S}^*$ are concatenated in the temporal order to generate a video summary.

\subsection{Implementation Detail}

\subsubsection{Deep feature computation.}
We uniformly extracted 5-second video segments in a temporal sliding window manner, where the window was shifted by 1 second. Each segment $V$ was re-sampled at 1 frame per second, so $V$ has five frames (\ie, $M=5$). The activations of VGG's ``fc7'' layer consists of 4,096 units. We set the unit size of the two fully connected layers to 1,000 and 300 respectively, which means our deep feature is a 300-dimensional vector. For the description sub-network, the fully-connected layers on top of the RNN have the same sizes as the video sub-network's. During the training, we fixed the network parameters of VGG and skip-thought, but those of the top two fully-connected layers for both video and description sub-networks were updated. We sampled 20 negative pairs for each positive pair to compute the contrastive loss. Our DNN was trained over the MSR-VTT dataset \cite{Xu2016}, which consists of 1M video clips annotated with 20 descriptions for each. We used Adam \cite{Kingma2015} to optimize the network parameters with the learning rate of $2^{-4}$ and trained for 4 epochs.

\subsubsection{Video summarization generation.}
Given an input video, we sampled 5-second video segments in the same way as the training of our DNN, and extracted a deep feature from each segment. We then minimize the objective function in Eq.~(\ref{eq:vidsum}) with cost-effective lazy forward selection \cite{Gygli2014,Leskovec2007}. We set the summary length $K$ to be roughly 15\% of the input video's length following \cite{Gygli2014}.

\section{Experiment}

To demonstrate the advantages of incorporating our deep features in video summarization, we evaluated and compared our approach with some baselines. We used the SumMe dataset \cite{Gygli2014} consisting of 25 videos for evaluation. As the videos in this dataset are either unedited or slightly edited, unimportant or redundant parts are left in the videos. The dataset includes videos with various contents. It also provides manually created video summaries for each video, with which we compare our summaries. We compute the  f-measure that evaluates agreement to reference video summaries using the code provided in \cite{Gygli2014}.

\subsection{Baselines}

We compared our video summaries with following several baselines as well as recent video summarization approaches:
(i) \textbf{Manually-created} video summaries are a powerful baseline that may be viewed as the upper bound for automatic approaches. The SumMe dataset provides at least 15 manually-created video summaries whose length is 15\% of the original video. We computed the average f-measure of each manually-created video summary with letting each of the rest manually-created video summaries as ground truth (\ie, if there are 20 manually-created video summaries, we compute 19 f-measures for each summary in a pairwise manner and calculate their average). We denote the summary with the highest f-measure among all manually-created video summaries by the best-human video summary. (ii) \textbf{Uniform sampling} (Uni.) is widely used baseline for video summarization evaluation. (iii) We also compare to video summaries generated in the same approach as ours except that VGG's ``fc7'' activations were used instead of our deep features, which is referred to as \textbf{VGG}-based video summary.
(iv) \textbf{Attention-based} video summary (Attn.) is a recently proposed video summarization approach using visual attention \cite{Ejaz2013}. (v) \textbf{Interestingness-based} video summary (Intr.) refers to a supervised approach \cite{Gygli2014}, where the weights of multiple objectives are optimized using the SumMe dataset.

\subsection{Results}
\begin{figure}[t!]
    \centering
    {\scriptsize Car Railcrossing} \\
    \includegraphics[width=0.9\textwidth]{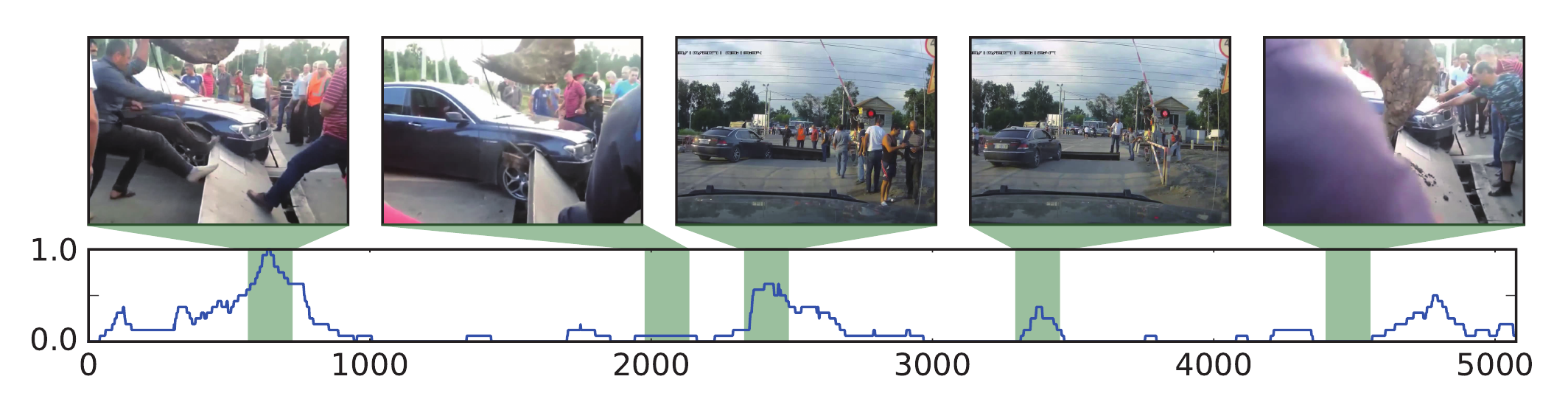} \\
    {\scriptsize Paluma Jump} \\
    \includegraphics[width=0.9\textwidth]{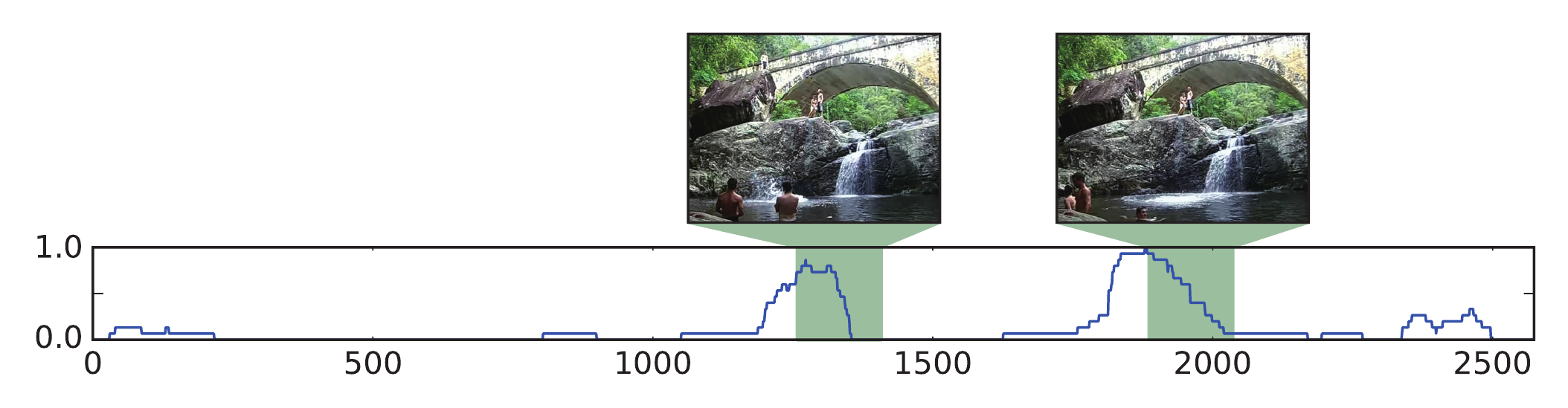} \\
    {\scriptsize Valparaiso Downhill} \\
    \includegraphics[width=0.9\textwidth]{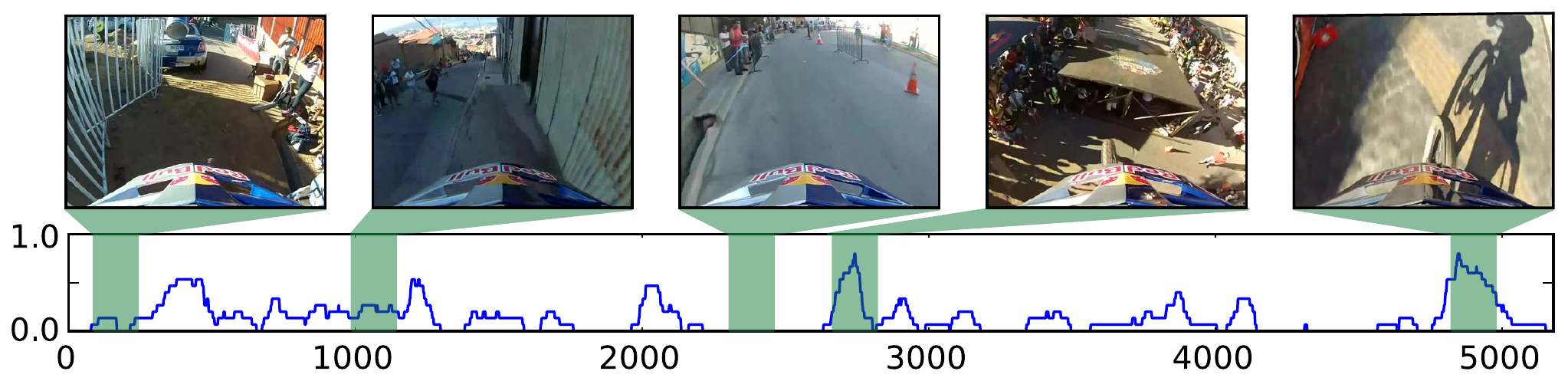}
    \caption{Segments selected by our approach. Keyframes of selected segments are shown. The green areas in the graphs indicate selected segments. The blue lines represents the ratio of annotators who selected the segment for their manually-created summary.}
    \label{fig:summary_wt_timeline} 
\end{figure}

Several examples of video summaries generated with our approach are shown in Fig.~\ref{fig:summary_wt_timeline}, along with ratio of annotators who agreed to include each video segments in their manually-created video summary. The peaks of the blue lines indicate that the corresponding video segments were frequently selected to create a video summary. These blue lines demonstrate that human annotators were consistent in some extent. Also we observe that the video segments selected by our approach (green areas) are correlated to the blue lines. This suggests that our approach is consistent with the human annotators.

\begin{table}[t!]
\centering
\caption{F-measures of manually-created video summaries and computational approaches (our approach and baselines, higher is better). Since there are multiple manually-created video summaries for each original video and thus multiple f-measures, we show their minimum, mean, and maximum. The best score among the computational approaches are highlighted.}
\label{tbl:summe_res}
\begin{tabular}{@{}r|ccc|ccccc@{}}
\toprule
\multicolumn{1}{l|}{}      & \multicolumn{3}{C{3cm}|}{Manually created}              & \multicolumn{5}{C{5cm}}{Computational approaches}                                                                              \\ \midrule
\multicolumn{1}{c|}{Video} & \multicolumn{1}{C{1cm}|}{Min.} & \multicolumn{1}{C{1cm}|}{Avg.} & Max. & \multicolumn{1}{C{1cm}|}{Uni.} & \multicolumn{1}{C{1cm}|}{VGG} & \multicolumn{1}{C{1cm}|}{Attn.} & \multicolumn{1}{C{1cm}|}{Intr.} & Ours \\ \midrule

Air Force One              & 0.185   & 0.332  & 0.457  & 0.060  & 0.239  & 0.215  & \textbf{0.318}  & 0.316  \\
Base Jumping               & 0.113   & 0.257  & 0.396  & \textbf{0.247}  & 0.062  & 0.194  & 0.121  & 0.077  \\
Bearpark Climbing          & 0.129   & 0.208  & 0.267  & 0.225  & 0.134  & \textbf{0.227}  & 0.118  & 0.178  \\
Bike Polo                  & 0.190   & 0.322  & 0.436  & 0.190  & 0.069  & 0.076  & \textbf{0.356}  & 0.235  \\
Bus in Rock Tunnel         & 0.126   & 0.198  & 0.270  & 0.114  & 0.120  & 0.112  & 0.135  & \textbf{0.151}  \\
Car Railcrossing           & 0.245   & 0.357  & 0.454  & 0.185  & 0.139  & 0.064  & \textbf{0.362}  & 0.328  \\
Cockpit Landing            & 0.110   & 0.279  & 0.366  & 0.103  & \textbf{0.190}  & 0.116  & 0.172  & 0.165  \\
Cooking                    & 0.273   & 0.379  & 0.496  & 0.076  & 0.285  & 0.118  & 0.321  & \textbf{0.329}  \\
Eiffel Tower               & 0.233   & 0.312  & 0.426  & 0.142  & 0.008  & 0.136  & \textbf{0.295}  & 0.174  \\
Excavators River Crossing  & 0.108   & 0.303  & 0.397  & 0.107  & 0.030  & 0.041  & \textbf{0.189}  & 0.134  \\
Fire Domino                & 0.170   & 0.394  & 0.517  & 0.103  & 0.124  & \textbf{0.252}  & 0.130  & 0.022  \\
Jumps                      & 0.214   & 0.483  & 0.569  & 0.054  & 0.000  & 0.243  & \textbf{0.427}  & 0.015  \\
Kids Playing in Leaves     & 0.141   & 0.289  & 0.416  & 0.051  & 0.243  & 0.084  & 0.089  & \textbf{0.278}  \\
Notre Dame                 & 0.179   & 0.231  & 0.287  & 0.156  & 0.136  & 0.138  & \textbf{0.235}  & 0.093  \\
Paintball                  & 0.145   & 0.399  & 0.503  & 0.071  & 0.270  & 0.281  & \textbf{0.320}  & 0.274  \\
Playing on Water Slide     & 0.139   & 0.195  & 0.284  & 0.075  & 0.092  & 0.124  & \textbf{0.200}  & 0.183  \\
Saving Dolphines           & 0.095   & 0.188  & 0.242  & 0.146  & 0.103  & \textbf{0.154}  & 0.145  & 0.121  \\
Scuba                      & 0.109   & 0.217  & 0.302  & 0.070  & 0.160  & \textbf{0.200}  & 0.184  & 0.154  \\
St Maarten Landing         & 0.365   & 0.496  & 0.606  & 0.152  & 0.153  & \textbf{0.419}  & 0.313  & 0.015  \\
Statue of Liberty          & 0.096   & 0.184  & 0.280  & 0.184  & 0.098  & 0.083  & \textbf{0.192}  & 0.143  \\
Uncut Evening Flight       & 0.206   & 0.350  & 0.421  & 0.074  & 0.168  & \textbf{0.299}  & 0.271  & 0.168  \\
Valparaiso Downhill        & 0.148   & 0.272  & 0.400  & 0.083  & 0.110  & 0.231  & 0.242  & \textbf{0.258}  \\
Car over Camera            & 0.214   & 0.346  & 0.418  & 0.245  & 0.048  & 0.201  & \textbf{0.372}  & 0.132  \\
Paluma Jump                & 0.346   & 0.509  & 0.642  & 0.058  & 0.056  & 0.028  & 0.181  & \textbf{0.428}  \\
playing ball               & 0.190   & 0.271  & 0.364  & 0.123  & 0.127  & 0.140  & 0.174  & \textbf{0.194}  \\ \midrule
Mean f-measure             & 0.179   & 0.311  & 0.409  & 0.124  & 0.127  & 0.167  & \textbf{0.234}  & 0.183  \\
Relative to human avg.     & 0.576   & 1.000  & 1.315  & 0.398  & 0.408  & 0.537  & \textbf{0.752}  & 0.588  \\
Relative to human max.     & 0.438   & 0.760  & 1.000  & 0.303  & 0.310  & 0.408  & \textbf{0.572}  & 0.447  \\ \bottomrule
\end{tabular}
\end{table}

The results of the quantitative evaluation are shown in the Table \ref{tbl:summe_res}. 
In this table, we report the minimum, average, and maximum f-measure scores of manually-created video summaries. Compared to VGG-based summary, ours significantly improved the scores. Our video summaries achieved 58.8\% of the average score of manually-created video summaries, while VGG-based got 40.8\%. This result demonstrates the advantage of our deep features for creating video summaries. 

One of the recent video summarization approaches, \ie, interestingness-based one \cite{Gygli2014}, got the highest score in this experiment. Note that the interestingness-based approach \cite{Gygli2014} uses a supervised technique, in which the mixture weights of various criteria in their objective function are optimized over the SumMe dataset. Our video summaries were generated using a relatively simple algorithm to extract a subset of segments; nevertheless, ours outperformed the interestingness-based for some videos, and even got a better mean f-measure score than attention-based.

Our approach got low scores, especially for short videos, such as ``Jumps'' and ``Fire Domino.'' Since we extract uniform length segments (5 second), in the case of short videos, our approach only extracts a few segments. This may result in a lower f-measure score. This limitation can be solved by extracting shorter video segments or using more sophisticated video segmentation like \cite{Sang2010,Gygli2014}.

We also observed that our approach got lower scores than others on the ``St Maarten Landing'' and ``Notre Dame,'' which are challenging because of long unimportant parts and diversity of content, respectively.
For ``St Maarten Landing,'' as our approach is unsupervised, it failed to exclude unimportant segments.
For ``Notre Dame,'' generating a summary is difficult because there are too many possible segments to be included in a summary.
While our summary shares small parts with manually created summaries,  it is a challenging example even for human annotators, which is shown in the low scores of manually-created video summaries.

\begin{figure}[t!]
    \centering
    {\scriptsize Cooking} \\
    \includegraphics[width=0.85\textwidth]{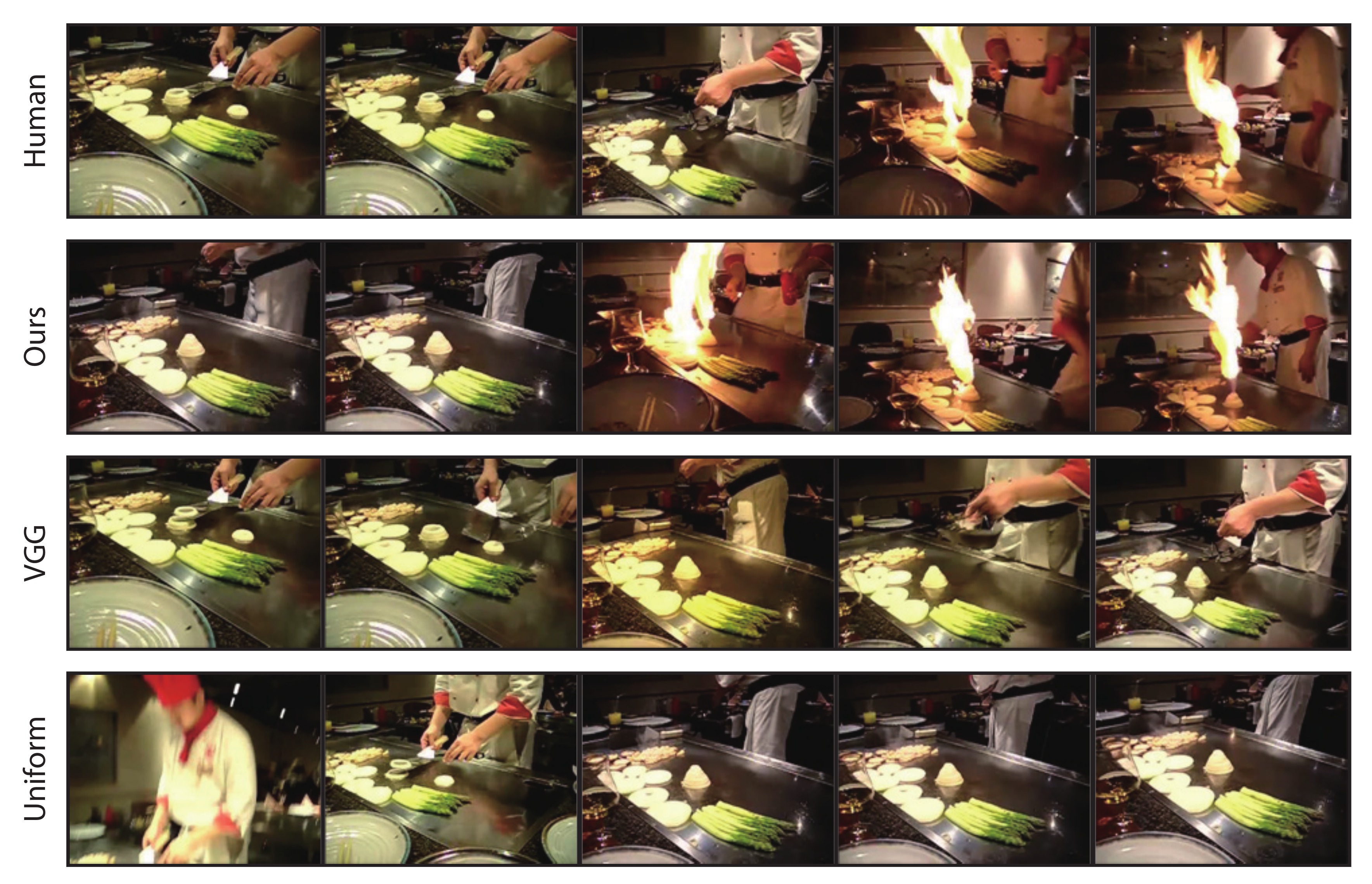} \\
    {\scriptsize Bear Climbing} \\
    \includegraphics[width=0.85\textwidth]{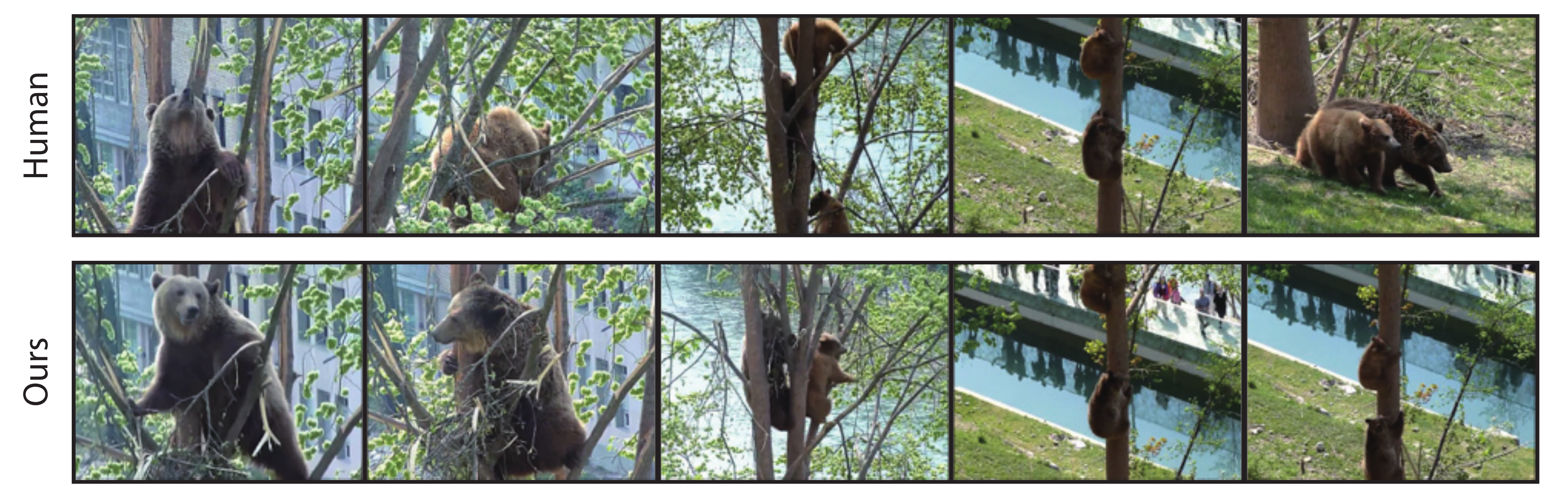} \\
    {\scriptsize Car over Camera} \\
    \includegraphics[width=0.85\textwidth]{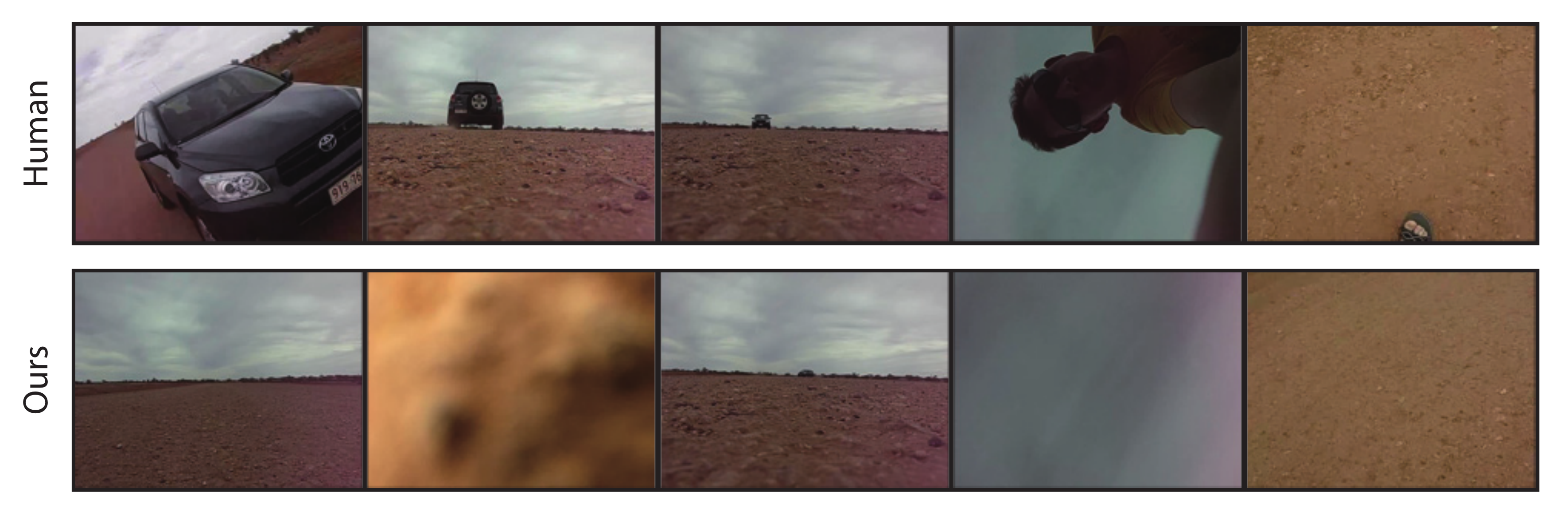}
    \caption{Uniformly sampled frames of summaries by different approaches. ``Human'' means the best-human video summary. The full results of ``Bear Climbing'' and ``Car over Camera'' are shown in the supplementary material.}
    \label{fig:summary_example} 
\end{figure}

Figure \ref{fig:summary_example} shows examples of video summaries created with our approach and baselines. The video ``Cooking'' shows a person cooking some vegetables while doing a performance. Ours and the best-human video summary include the same scene of the performance with fire, while others do not. On the other hand, ours extracts unimportant segments from the video ``Car over Camera.'' The original video is highly redundant with static scenes just showing the ground or the sky, and such scenes make up large clusters in the semantic space even if they are unimportant. As our approach extracts representatives from each cluster, a video with lengthy unimportant parts resulted in a poor video summaries. We believe that this problem can be avoided by using visual cues such as interestingness \cite{Gygli2013} and objectiveness \cite{Alexe2010}.

\section{Conclusion}
In this work, we proposed to learn semantic deep features for video summarization and a video summarization approach that extracts a video summary based on the representativeness in the semantic feature space.
For deep feature learning, we designed a DNN with two sub-networks for videos and descriptions, which are jointly trained using the contrastive loss.
We observed that learned features extracted from videos with similar content make clusters in the semantic space.
In our approach, the input video is represented by deep features in the semantic space, and segments corresponding to cluster centers are extracted to generate a video summary.
By comparing our summaries to manually created summaries, we shown that the advantage of incorporating our deep features in a video summarization technique.
Furthermore, our results even outperformed the worst human created summaries.
We expect that the quality of video summaries will be improved by incorporating video segmentation methods.
Moreover, our objective function can be extended by considering other criteria used in the area of video summarization, such as interestingness and temporal uniformity. 


\vspace{3mm}
\noindent {\bf Acknowledgement}. This work is partly supported by JSPS KAKENHI No. 16K16086.

\bibliographystyle{splncs}
\bibliography{egbib}

\begin{thebibliography}{10}

\bibitem{youtube_stat}
{YouTube.com}:
\newblock Statistics--{YouTube}.
\newblock \url{https://www.youtube.com/yt/press/en-GB/statistics.html} (2016)

\bibitem{Gong2000}
Gong, Y., Liu, X.:
\newblock Video summarization using singular value decomposition.
\newblock In: Proc. IEEE Computer Society Conf. Computer Vision and Pattern
  Recognition (CVPR). (2000)  174--180

\bibitem{Gong2014}
Gong, B., Chao, W.L., Grauman, K., Sha, F.:
\newblock Diverse sequential subset selection for supervised video
  summarization.
\newblock In: Proc. Advances in Neural Information Processing Systems (NIPS).
  (2014)  2069--2077

\bibitem{Zhao2014}
Zhao, B., Xing, E.P.:
\newblock Quasi real-time summarization for consumer videos.
\newblock In: Proc. IEEE Computer Society Conf. Computer Vision and Pattern
  Recognition (CVPR). (2014)  2513--2520

\bibitem{Lowe2004}
Lowe, D.G.:
\newblock {Distinctive image features from scale invariant keypoints}.
\newblock Int. Journal of Computer Vision \textbf{60} (2004)  91--11020042

\bibitem{Dalal2005}
Dalal, N., Triggs, B.:
\newblock Histograms of oriented gradients for human detection.
\newblock In: Proc. IEEE Computer Society Conf. Computer Vision and Pattern
  Recognition (CVPR). (2005)  886--893

\bibitem{Yao2015}
Yao, L., Ballas, N., Larochelle, H., Courville, A.:
\newblock Describing videos by exploiting temporal structure.
\newblock In: Proc. IEEE Int. Conf. Computer Vision (ICCV). (2015)  4507--4515

\bibitem{Donahue2014}
Donahue, J., Jia, Y., Vinyals, O., Hoffman, J., Zhang, N., Tzeng, E., Darrell,
  T.:
\newblock {DeCAF: A} deep convolutional activation feature for generic visual
  recognition.
\newblock In: Proc. Int. Conf. Machine Learning (ICML). Volume~32. (2014)
  647--655

\bibitem{Xu2016}
Xu, J., Mei, T., Yao, T., Rui, Y.:
\newblock {MSR-VTT: A} large video description dataset for bridging video and
  language.
\newblock In: Proc. IEEE Computer Society Conf. Computer Vision and Pattern
  Recognition (CVPR). (2016)  5288--5296

\bibitem{Simonyan2015}
Simonyan, K., Zisserman, A.:
\newblock Very deep convolutional networks for large-scale image recoginition.
\newblock In: Proc. Int. Conf. Learning Representations (ICLR). (2015)  14
  pages

\bibitem{Babaguchi2004}
Babaguchi, N., Kawai, Y., Ogura, T., Kitahashi, T.:
\newblock Personalized abstraction of broadcasted {American} football video by
  highlight selection.
\newblock IEEE Trans. Multimedia \textbf{6} (2004)  575--586

\bibitem{Sang2010}
Sang, J., Xu, C.:
\newblock Character-based movie summarization.
\newblock In: Proc. ACM Int. Conf. Multimedia (MM). (2010)  855--858

\bibitem{Evangelopoulos2013}
Evangelopoulos, G., Zlatintsi, A., Potamianos, A., Maragos, P., Rapantzikos,
  K., Skoumas, G., Avrithis, Y.:
\newblock Multimodal saliency and fusion for movie summarization based on
  aural, visual, and textual attention.
\newblock IEEE Trans. Multimedia \textbf{15} (2013)  1553--1568

\bibitem{Lu2013}
Lu, Z., Grauman, K.:
\newblock Story-driven summarization for egocentric video.
\newblock In: Proc. IEEE Computer Society Conf. Computer Vision and Pattern
  Recognition (CVPR). (2013)  2714--2721

\bibitem{Potapov2014}
Potapov, D., Douze, M., Harchaoui, Z., Schmid, C.:
\newblock Category-specific video summarization.
\newblock In: Proc. European Conf. Computer Vision (ECCV). (2014)  540--555

\bibitem{Yang2015}
Yang, H., Wang, B., Lin, S., Wipf, D., Guo, M., Guo, B.:
\newblock {Unsupervised extraction of video highlights via robust recurrent
  auto-encoders}.
\newblock In: Proc. IEEE Int. Conf. Computer Vision (ICCV). (2015)  4633--4641

\bibitem{Xu2015}
Xu, J., Mukherjee, L., Li, Y., Warner, J., Rehg, J.M., Singh, V.:
\newblock Gaze-enabled egocentric video summarization via constrained
  submodular maximization.
\newblock In: Proc. IEEE Computer Society Conf. Computer Vision and Pattern
  Recognition (CVPR). (2015)  2235--2244

\bibitem{Tschiatschek2014}
Tschiatschek, S., Iyer, R.K., Wei, H., Bilmes, J.A.:
\newblock Learning mixtures of submodular functions for image collection
  summarization.
\newblock In: Proc. Advances in Neural Information Processing Systems (NIPS).
  (2014)  1413--1421

\bibitem{Gygli2014}
Gygli, M., Grabner, H., Riemenschneider, H., {van Gool}, L.:
\newblock Creating summaries from user videos.
\newblock In: Proc. European Conf. Computer Vision (ECCV). (2014)  505--520

\bibitem{Gygli2015}
Gygli, M., Grabner, H., {van Gool}, L.:
\newblock Video summarization by learning submodular mixtures of objectives.
\newblock In: Proc. IEEE Computer Society Conf. Computer Vision and Pattern
  Recognition (CVPR). (2015)  3090--3098

\bibitem{Song2015}
Song, Y., Vallmitjana, J., Stent, A., Jaimes, A.:
\newblock {TVSum: Summarizing web videos using titles}.
\newblock In: Proc. IEEE Computer Society Conf. Computer Vision and Pattern
  Recognition (CVPR). (2015)  5179--5187

\bibitem{Khosla2013}
Khosla, A., Hamid, R., Lin, C.j., Sundaresan, N.:
\newblock Large-scale video summarization using web-image priors.
\newblock In: Proc. IEEE Computer Society Conf. Computer Vision and Pattern
  Recognition (CVPR). (2013)  2698--2705

\bibitem{Chu2015}
Chu, W.S., Jaimes, A.:
\newblock Video co-summarization: Video summarization by visual co-occurrence.
\newblock In: Proc. IEEE Computer Society Conf. Computer Vision and Pattern
  Recognition (CVPR). (2015)  3584--3592

\bibitem{Wang2015}
Wang, X., Gupta, A.:
\newblock Unsupervised learning of visual representations using videos.
\newblock In: Proc. IEEE Int. Conf. Computer Vision (ICCV). (2015)  2794--2802

\bibitem{Frome2013}
Frome, A., Corrado, G., Shlens, J.:
\newblock {DeViSE: A} deep visual-semantic embedding model.
\newblock In: Proc. Advances in Neural Information Processing Systems (NIPS).
  (2013)  2121--2129

\bibitem{Chopra2005}
Chopra, S., Hadsell, R., LeCun, Y.:
\newblock Learning a similarity metric discriminatively, with application to
  face verification.
\newblock In: Proc. IEEE Computer Society Conf. Computer Vision and Pattern
  Recognition (CVPR). (2005)  539--546

\bibitem{Lin2015}
Lin, T.Y., Belongie, S., Hays, J.:
\newblock Learning deep representations for ground-to-aerial geolocalization.
\newblock In: Proc. IEEE Computer Society Conf. Computer Vision and Pattern
  Recognition (CVPR). (2015)  5007--5015

\bibitem{Kiros2015}
Kiros, R., Zhu, Y., Salakhutdinov, R.R., Zemel, R., Urtasun, R., Torralba, A.,
  Fidler, S.:
\newblock Skip-thought vectors.
\newblock In: Proc. Advances in Neural Information Processing Systems (NIPS).
  (2015)  3276--3284

\bibitem{VanderMaaten2008}
{van} {der}~Maaten, L., Hinton, G.E.:
\newblock Visualizing high-dimensional data using t-{SNE}.
\newblock Journal of Machine Learning Research \textbf{9} (2008)  2579--2605

\bibitem{Daniel}
DeMenthon, D., Kobla, V., Doermann, D.:
\newblock Video summarization by curve simplification.
\newblock In: Proc. ACM Int. Conf. Multimedia (MM). (1998)  211--218

\bibitem{Kingma2015}
Kingma, D., Ba, J.:
\newblock Adam: A method for stochastic optimization.
\newblock In: Proc. Int. Conf. Learning Representations (ICLR). (2015)  11
  pages

\bibitem{Leskovec2007}
Leskovec, J., Krause, A., Guestrin, C., Faloutsos, C., VanBriesen, J., Glance,
  N.:
\newblock Cost-effective outbreak detection in networks.
\newblock Proc. ACM SIGKDD Int. Conf. Knowledge Discovery and Data Mining (KDD)
  (2007)  420--429

\bibitem{Ejaz2013}
Ejaz, N., Mehmood, I., {Wook Baik}, S.:
\newblock Efficient visual attention based framework for extracting key frames
  from videos.
\newblock Signal Processing: Image Communication \textbf{28} (2013)  34--44

\bibitem{Gygli2013}
Gygli, M., Grabner, H., Riemenschneider, H., Nater, F., Gool, L.V.:
\newblock The interestingness of images.
\newblock In: IEEE Int. Conf. Computer Vision (ICCV). (2013)  1633--1640

\bibitem{Alexe2010}
Alexe, B., Deselaers, T., Ferrari, V.:
\newblock What is an object?
\newblock In: Proc. IEEE Computer Society Conf. Computer Vision and Pattern
  Recognition (CVPR). (2010)  73--80

\end{thebibliography}



\end{document}